\documentclass[runningheads]{llncs}
\usepackage{graphicx}
\usepackage{enumitem}
\setdescription{leftmargin=0pt}
\usepackage{graphicx}
\usepackage{rotating}
\usepackage{algorithm2e}
\usepackage{float}
\restylefloat{table}
\usepackage{algpseudocode}
\usepackage{amsmath}
\usepackage{amssymb}
\usepackage{hyperref}
\usepackage{amsfonts}
\usepackage{stmaryrd}
\usepackage{fixcmex}

\DeclareFontFamily{OMS}{cmtt}{\hyphenchar\font45 }
\DeclareFontShape{OMS}{cmtt}{m}{n}{<->ssub*cmsy/m/n}{}

\begin{document}
\title{Evaluating Class Membership Relations in Knowledge Graphs using Large Language Models}
\titlerunning{Evaluating Class Membership Relations in Knowledge Graphs using LLMs}
%
\author{Bradley P. Allen\orcidID{0000-0003-0216-3930} \and
Paul T. Groth\orcidID{0000-0003-0183-6910}}
\authorrunning{B.P. Allen and P.T. Groth}
%
\institute{University of Amsterdam, Amsterdam, The Netherlands \\
\email{\{b.p.allen,p.t.groth\}@uva.nl}}
\maketitle              
\begin{abstract}
A backbone of knowledge graphs are their class membership relations, which assign entities to a given class. As part of the knowledge engineering process, we propose a new method for evaluating the quality of these relations by processing descriptions of a given entity and class using a zero-shot chain-of-thought classifier that uses a natural language intensional definition of a class. We evaluate the method using two publicly available knowledge graphs, Wikidata and CaLiGraph, and 7 large language models. Using the gpt-4-0125-preview large language model, the method's classification performance achieves a macro-averaged F1-score of 0.830 on data from Wikidata and 0.893 on data from CaLiGraph. Moreover, a manual analysis of the classification errors shows that 40.9\% of errors were due to the knowledge graphs, with 16.0\% due to missing relations and 24.9\% due to incorrectly asserted relations. These results show how large language models can assist knowledge engineers in the process of knowledge graph refinement. The code and data are available on Github\footnote{\url{https://github.com/bradleypallen/evaluating-kg-class-memberships-using-llms}}.
\end{abstract}

\section{Introduction}
Knowledge graphs (KGs) have become a key technology in many applications in industry and academia \cite{10.1145/3447772}. This has brought attention to the area of KG refinement \cite{paulheim2017knowledge}, for which a main goal is ensuring that the knowledge captured in KGs is as complete and correct as possible. This is a challenge, given that large-scale KGs composed of contributions from multiple sources of knowledge often contain incomplete, misaligned, and inaccurate information \cite{shenoy2022study,piscopo2019we}. At the same time, as part of the knowledge engineering process, direct manual evaluation of KG quality by human reviewers to detect and remediate these problems is expensive \cite{xue2022knowledge,hofer2023construction}. 

The recent emergence of large language models (LLMs) has inspired work towards understanding how LLMs can applied to knowledge graph construction. To date, much of this work has centered on the use of LLMs for knowledge graph completion \cite{zhang2023using} and the evaluation of provenance \cite{amaral2022prove} and correctness \cite{syed2018factcheck} in a knowledge graph. In this paper, we describe work on using LLMs to evaluate {\em class membership relations} in a KG. Class membership relations are important because they are a principal way in which knowledge graphs represent classification schemes. Classification schemes are a major consideration in many knowledge engineering efforts, often with significant implications for social policy and scientific consensus \cite{bowker2000sorting}.

We present an approach to evaluate class membership relations by using an LLM to define a zero-shot chain-of-thought (CoT) \cite{kojima2022large,wei2022chain} classifier that takes natural language descriptions of an entity and a class in a given KG, and predicts whether or not the entity is an instance of the class, providing a natural language rationale for the prediction. The motivation for this approach is to leverage an LLM's capabilities for natural language processing to allow knowledge engineers to use intensional knowledge expressed in natural language by domain experts directly, as opposed to having to first transform it into a symbolic knowledge representation, and apply it to determining if that knowledge is accurately reflected in a given knowledge graph. 

\section{Related work}
\begin{description}
    \item[Using LLMs for knowledge engineering tasks]
    Beyond uses for KG refinement, LLMs are beginning to be applied to other tasks in the engineering of knowledge graphs. In \cite{allen2023knowledge}, two scenarios for the use of LLMs in knowledge engineering are described: creating hybrid neurosymbolic knowledge systems and enabling knowledge engineering in natural language. Pan et al \cite{pan2023unifying} describe three categories of LLM/KG hybrids: KG-enhanced LLMs, LLM-augmented KGs, and synergized LLMs + KGs. Specific examples of LLM augmentation of KGs include the use of LLMs for KG completion \cite{zhang2023using,alivanistos2022prompting} and for ontology engineering \cite{mateiu2023ontology}. We view our work as an example of an LLM-augmented KG approach that performs knowledge engineering using intensional knowledge expressed in natural language to develop classifiers; classification is a well-known instance of an analytic knowledge task as defined in the CommonKADS taxonomy of knowledge-intensive task types \cite{schreiber2000knowledge}.
    \item[KG refinement] 
    Knowledge graph refinement is defined by Paulheim  \cite{paulheim2017knowledge} as the process of improving an existing KG by adding missing knowledge or identifying and removing errors. 
    KG refinement has been implemented using manual, statistical, rule-based and hybrid methods \cite{xue2022knowledge,hofer2023construction}. Interactive solutions to aid human reviewers have been developed, including tools for crowdsourcing KG quality assessment \cite{kontokostas2013triplecheckmate}, fact-checking triples using textual evidence \cite{syed2018factcheck}, ontology repair using description logic reasoners \cite{lambrix2023completing}, and sampling techniques to better focus manual reviewers' attention \cite{gao2019efficient}. Our work builds on these results by creating classifiers that can be used to alert a knowledge engineer to misalignments between natural language definitions of a class and elements of the class's extension in a given KG.
    \item[Automated fact checking]
    A recent survey \cite{guo2022survey} provides a useful overview of the large amount of methods for fact checking. The work most related to ours is that of Atanasova et al. \cite{atanasova2020generating} on justification production using language models of the BERT family. That work focuses on the fact checking applied to claims expressed as natural language statements; in contrast, our methods admit of the combination of both serialized RDF statements and natural language descriptions as input for both justification production and verdict prediction.
\end{description}

\section{Preliminaries}
To precisely specify the integration between KGs and LLMs in our experiments, we now introduce a formalization of a neurosymbolic workflow \cite{ekaputra2023describing} for entity classification.
\begin{description}
\item[Language models] Let \( \mathcal{T} \) be the set of sequences of tokens \( T_i = t_1, t_2, \ldots, t_n \) such that \( t_i \) is a token in a predefined vocabulary \( V \). Given a \textit{corpus} \( \mathcal{C} \subseteq \mathcal{T} \), a \textit{language model} \( \mathcal{L_C} \) is a probabilistic model trained on a sample of \( \mathcal{C} \) that defines a distribution over sequences of tokens.  
\begin{equation}
\mathcal{L_C}(T_i) = p(t_1, t_2, \ldots, t_n)         
\end{equation}
is an estimate of the probability of a sequence \( T_i \), given a corpus \( \mathcal{C} \).
A \textit{prompt} \( P = (T, F) \) is a pair of a sequence of tokens \( T \) and an set of \textit{free} tokens \( F \subseteq \{ f_1, f_2, \ldots, f_n \} \). A \textit{substitution} \( \theta \) with respect to a prompt \( P \) is a set of pairs \( ( f_i, T_i ) \) such that \( f_i \in F \) and \( T_i \in \mathcal{T} \). An \textit{instantiation} \( \texttt{instantiate}(P, \theta) \) is a prompt \( P' \) such that \( \forall (f_i, T_i) \in \theta \) every occurrence of \( f_i \) in \( P \) is replaced with \( f_i \).
Given a prompt \( P \), the goal of a language model \( \mathcal{L_C} \) is to generate a sequence of tokens that maximizes the conditional probability under \( \mathcal{L_C} \).
\begin{equation}
T_{\text{out}} = \arg \max_{T} \mathcal{L_C}(T | P)        
\end{equation}
is the output sequence generated by the language model, conditioned on \( P \).
\item[Knowledge graphs] Following \cite{angles2020mapping}, we use the RDF data model to describe knowledge graphs. Let \( I \) be an infinite set of IRIs (Internationalized Resource Identifiers \cite{durst2005internationalized}), \( B \) be an infinite set of blank nodes \cite{hogan2014everything}, and \( L \) an infinite set of literals \cite{beek2018literally}. A \textit{knowledge graph} \( G \) is a set of \textit{triples} \( \{(s, p, o) \mid s \in S, p \in P, o \in O \} \), where \( S \subset I \cup B \) is the set of \textit{subjects} in \( G \), \( P \subset I \) is the set of \textit{properties} in \( G \), and \( O \subset I \cup B \cup L \) is the set of \textit{objects} in \( G \).
Let \( \texttt{instanceOf}, \texttt{subClassOf}, \texttt{label} \in P \) denote an instance-of relation, a subclass-of relation, and a label property in \( G \), respectively.
A \textit{class} \( c \in I \cup B \) is an entity 
that represents a set of entities sharing common properties and relationships in $G$.
Let
\begin{equation}
\texttt{ext}(c) = \bigcup_{i \in \mathop \mathbb{N}} \texttt{ext}_i(c)        
\end{equation}
be the \textit{extension} of a class \( c \), where
\begin{equation}
    \begin{aligned}
\texttt{ext}_0(c) = \{ e \mid \exists (e, \texttt{instanceOf}, c) \in G \}        
    \end{aligned}
\end{equation}
\begin{equation}
\begin{aligned}
\texttt{ext}_{i+1}(c) =  & \texttt{ext}_i(c) \cup \{ e \mid e \in \texttt{ext}(c') \land  \exists (c', \texttt{subClassOf}, c) \in G \}
\end{aligned}
\end{equation}
\item[Zero-shot chain-of-thought entity classifiers] Given the definitions above, we now proceed to show how to construct classifiers that prompt LLMs with intensional definitions of classes in natural language to classify entities in a knowledge graph. For any entity \( e \in I \cup B \), let \( Ge = \{ (s, p, o) \in G \mid s = e \lor o = e \} \) be the \textit{neighborhood} of $e$. Let \( T_{label(e)} = \{ o \mid \exists (e, \texttt{label}, o) \in G \) \}. A \textit{serialization} \( T_G \) of a knowledge graph \( G \) is a sequence of tokens \( T \) that represents the triples in \( G \) using a structured formal language (e.g. RDF). For any entity \( e \in E \), let \( T_{Ge} \) be the serialization of $Ge$. 
A \textit{verbalization} \( T_e \) of an entity \( e \) is a sequence of tokens \( T \) that represents a description of $e$ in natural language. 
Given an language model $\mathcal{L_C}$, we define a function \texttt{classify} as follows:
\begin{equation}
    \begin{aligned}
( T_R, T_\mathbb{B} ) = \texttt{classify}(c, e)      
    \end{aligned}
\end{equation}
where $T_R$ is a sequence of tokens that represents a rationale for a classification decision, and $T_\mathbb{B} \in \{ \texttt{positive}, \texttt{negative} \}$ are tokens that represent classification decisions, i.e., whether or not $e \in \texttt{ext}(c)$, respectively.
We instantiate $T_R$ and $T_\mathbb{B}$ as follows:
\begin{equation}
    \begin{aligned}
T_R = \arg \max_{T} \mathcal{L_C}(T | \texttt{instantiate}(P_{rationale\_generation},\theta_0))        
    \end{aligned}
\end{equation}
\begin{equation}
    \begin{aligned}
T_\mathbb{B} = \arg \max_{T} \mathcal{L_C}(T | \texttt{instantiate}(P_{answer\_generation},\theta_1))        
    \end{aligned}
\end{equation}
\begin{equation}
\begin{aligned}
\theta_0 = \{ & (\texttt{\{label\}}, T_{label(c)}),(\texttt{\{definition\}}, T_{c}), \\
& (\texttt{\{entity\}}, T_{label(e)}),(\texttt{\{description\}}, T_{e}) \} 
\end{aligned}
\end{equation}
\begin{equation}
    \begin{aligned}
\theta_1 = \theta_0 \cup \{(\texttt{\{rationale\}}, T_{R_e})\}        
    \end{aligned}
\end{equation}
given two prompt templates $P_{rationale\_generation}$ and $P_{answer\_generation}$. The specific prompt templates used in the experiments were manually authored and iteratively refined between June 2023 and October 2023. 
Figure \ref{fig:zero_shot_cot_classifier} shows an example of such a classifier instantiated for a class and entity in the CaLiGraph KG. 
\begin{sidewaysfigure}
\vspace*{\fill}
\centering
\includegraphics[width=0.9\textheight]{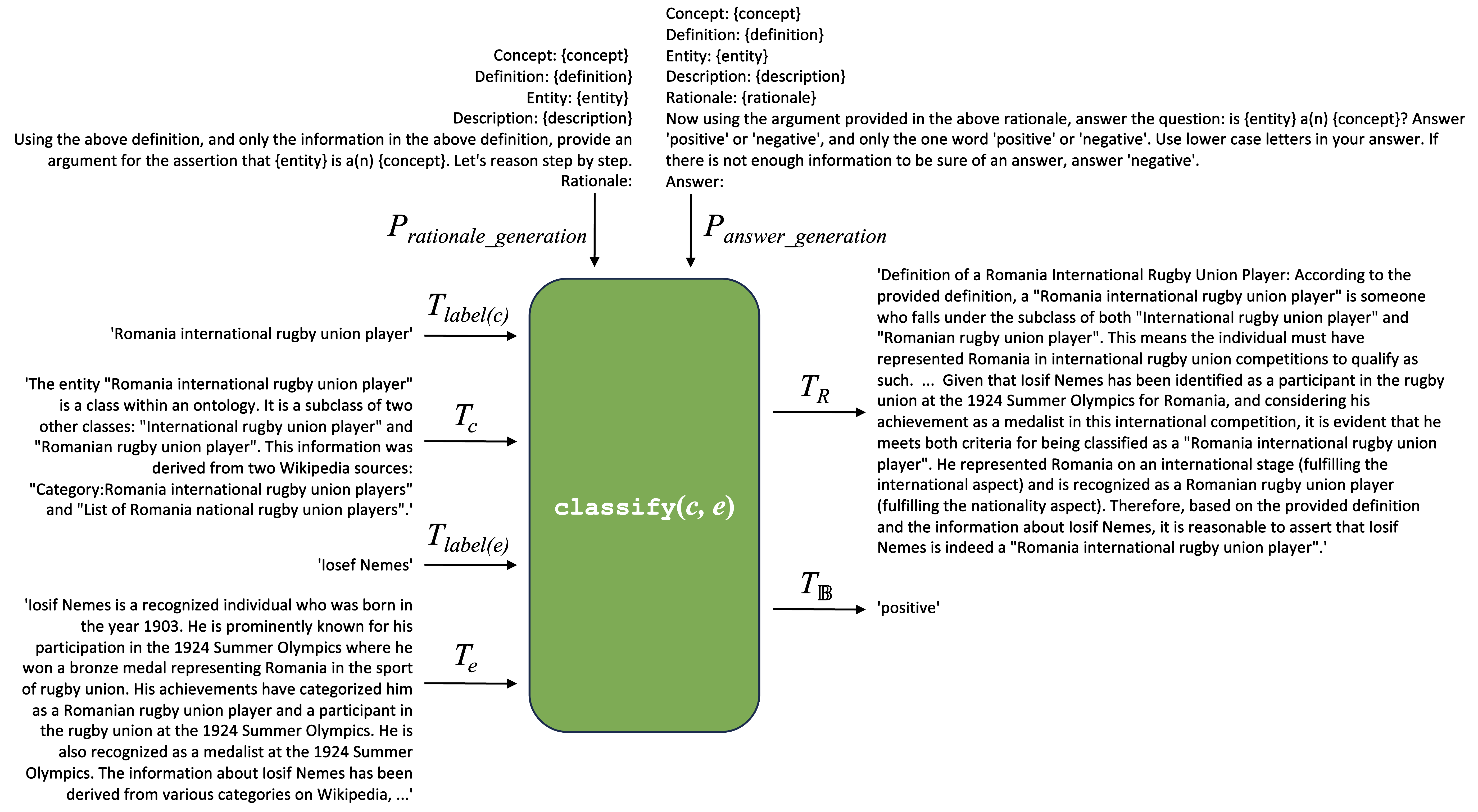}
\parbox{0.9\textwidth}{
\caption{A zero-shot chain-of-thought classifier applied to the class \texttt{clgo:Romania\_international\_rugby\_union\_player} and the entity \texttt{clgr:Iosif\_Nemes} from the CaLiGraph knowledge graph \cite{heist2021caligraph}.}
}
\label{fig:zero_shot_cot_classifier}
\end{sidewaysfigure}
\end{description}

\section{Experiments}
To understand the potential of classifiers built using the above approach for the problem of KG refinement, we conducted experiments to explore two research questions:
\begin{description}
\item[$Q_1$: Can the classifiers exhibit good alignment with KGs?] Much of the work on LLM/KG synergy to date is predicated on the idea that KGs, as curated sources of knowledge, can be used to address gaps in the knowledge obtainable from LLMs, or mitigate the problem of hallucination by grounding LLMs. This makes assumptions about the degree of alignment between LLMs and KGs, hence this question aims to measure this alignment.
\item[$Q_2$: Can the classifiers detect missing or incorrect relations?] Our main goal is to generate classifications based on intensional class definitions in with natural language rationales to help guide human reviewers to areas where KGs may be incomplete or incorrect. Any such approach must demonstrate the ability to do so across multiple knowledge graphs and classes.
\end{description}
The experiments to address these questions were implemented as follows:
\begin{description}
\item[Knowledge graphs] Two publicly available knowledge graphs were used to construct evaluation datasets: Wikidata \cite{vrandevcic2014wikidata} and CaLiGraph \cite{heist2021caligraph,heist2021information,heist2022transformer}. The two KGs represent distinct approaches to KG construction. Wikidata is the result of the crowd-sourced contribution of factual statements by thousands of human contributors and automated processes working independently, yielding relatively diverse approaches to modeling concepts and entities, and is loosely coupled with and derived from information in Wikipedia. CaLiGraph is the result of the automated extraction of terminology and assertions from Wikipedia and DBPedia pages, and as such is more consistent in how it models concepts and entities than is Wikidata. 
\item[Data sets] 
We randomly sampled 20 classes from Wikidata and 19 from CaLiGraph using SPARQL queries, including their super-classes. For each, 20 entities were selected as positive examples, and up to 20 entities from the set difference of the extensions of the class and one of its superclasses as negative examples (in some of the sampled classes the cardinality of the set difference was less than 20). 
\item[Language models] We evaluated seven large language models accessible using services provided by OpenAI and Hugging Face: OpenAI's gpt-4-0125-preview and gpt-3.5-turbo, Google's gemma-7b-it and gemma-2b-it, Mistral AI's Mixtral-8x7B-Instruct-v0.1 and Mistral-7B-Instruct-v0.2, and Meta's Llama-2-70b-chat-hf. For all experiments, temperatures were set to a value of 0.1. 
\item[Classifiers]
We use the definitions provided above to instantiate a classifier for each class in the datasets. For each class $c$ and entity $e$, we obtained natural language descriptions to use as $T_c$ and $T_e$ arguments for the classifier. For Wikidata, we retrieved natural language summaries of the class or entity from its associated Wikipedia page. For CaLiGraph, we used gpt-4-1106-preview to generate RDF verbalizations to serve as $T_c$ and $T_e$, given inputs of $T_{G_c}$ and $T_{G_e}$ obtained as the TSV serialization of the triples returned from SPARQL DESCRIBE queries for $c$ and $e$ with LIMIT = 20.
\item[Evaluation procedure] Experimental runs were then conducted by applying classifiers to each class/entity pair for a given class in each of the two knowledge graphs, generating for each class a confusion matrix based on the resulting set of classifications, from which performance metrics are computed. 
Algorithm \ref{alg:experiment} describes this procedure in pseudo-code. Evaluations whose statistics are reported below were conducted during the period from 24 February 2024 to 27 February 2024. Costs incurred through calls to language model APIs during this period totalled around \$225 USD.
\begin{algorithm}[!ht]
\SetKwFunction{Classify}{classify}
\SetKwFunction{Ext}{ext}
    \SetKwInOut{Input}{input}
    \SetKwInOut{Output}{output}
    \Input{a pair of classes $c, d$ from $G \mid (c, \texttt{subClassOf}, d) \in G$}
    \Output{a confusion matrix $M$}
    \BlankLine
    ($TP$, $FP$, $TN$, $FN$) $\leftarrow$ (0, 0, 0, 0)\;\
    $E^+$ $\leftarrow$ a sample from \Ext{$c$}\;\
    $E^-$ $\leftarrow$ \text{a sample from} \Ext{$d$} $\setminus$ \Ext{$c$}\;
    \ForEach{$e \in E^+$}{
        $( T_R, T_\mathbb{B} )$ $\leftarrow$ \Classify{$c,e$}\;\
        \lIf{$T_\mathbb{B} = \texttt{positive}$} {$TP \leftarrow TP + 1$}\
        \lElse{$FP \leftarrow FP + 1$}\
    }
    \ForEach{$e \in E^-$}{
        $( T_R, T_\mathbb{B} )$ $\leftarrow$ \Classify{$c,e$}\;\
        \lIf{$T_\mathbb{B} = \texttt{negative}$} {$TN \leftarrow TN + 1$}\
        \lElse{$FN \leftarrow FN + 1$}\
    }
    $M \leftarrow [ [ TP, FP ], [ FN, TN ] ]$\;
    \BlankLine
    \caption{Evaluation procedure}
    \label{alg:experiment}
\end{algorithm}
\end{description}

\section{Findings}
We summarize below the finding obtained from our evaluations. Detailed results can be found in our aforementioned Github repository.  

Classifier performance (assuming the KG as ground truth) of the seven closed- and open-source LLMs is shown in Table \ref{tab:classifier-llm-performance}. The performance as shown supports the following finding: \textbf{classifiers can exhibit good alignment with KGs} ($Q_1$). As evidenced by Cohen's $\kappa$ values, one LLM was in moderate agreement with Wikidata, and four were in moderate agreement with CaLiGraph.
\begin{table}[!ht]
\centering
\scriptsize
\begin{tabular}{p{1.7cm}|p{4cm}|p{1cm}|p{1cm}|p{1cm}|p{1cm}}
KG & LLM & ACC & AUC & F1 & $\kappa$ \\ \hline
Wikidata & gpt-4-0125-preview & \textbf{0.830} & \textbf{0.830} & \textbf{0.823} & \textbf{0.660} \\ 
 & gemma-7b-it & 0.726 & 0.727 & 0.705 & 0.454 \\ 
 & Mixtral-8x7B-Instruct-v0.1 & 0.697 & 0.696 & 0.654 & 0.393 \\ 
 & Mistral-7B-Instruct-v0.2 & 0.671 & 0.671 & 0.620 & 0.342 \\ 
 & gemma-2b-it & 0.674 & 0.670 & 0.629 & 0.330 \\ 
 & gpt-3.5-turbo & 0.627 & 0.627 & 0.547 & 0.255 \\ 
 & Llama-2-70b-chat-hf & 0.631 & 0.616 & 0.569 & 0.239 \\ \hline 
CaLiGraph & gpt-4-0125-preview & \textbf{0.900} & \textbf{0.893} & \textbf{0.889} & \textbf{0.788} \\ 
 & Mixtral-8x7B-Instruct-v0.1 & 0.893 & 0.884 & 0.874 & 0.767 \\ 
 & gpt-3.5-turbo & 0.842 & 0.833 & 0.815 & 0.665 \\ 
 & Mistral-7B-Instruct-v0.2 & 0.812 & 0.803 & 0.779 & 0.605 \\ 
 & gemma-7b-it & 0.783 & 0.774 & 0.750 & 0.547 \\ 
 & Llama-2-70b-chat-hf & 0.637 & 0.625 & 0.558 & 0.252 \\ 
 & gemma-2b-it & 0.563 & 0.543 & 0.422 & 0.090 \\ 
\end{tabular}
\vspace*{2mm}
\caption{Classifier performance by LLM.}
\label{tab:classifier-llm-performance}
\end{table}

Table \ref{tab:summary_error_analysis} shows the results of an error analysis of the evaluation results for the highest-performing classifier (using gpt-4-0125-preview). It was conducted by having one of the authors manually annotate each classification error with their own classification decision, based on the information in the provided descriptions. This human judgment was then compared with that of the KG and classifier using the pairwise Cohen's $\kappa$ value as a measure of inter-annotator agreement. In cases where where Wikidata and the classifier using gpt-4-0125-preview disagreed, the human showed fair agreement with Wikidata and no agreement with the classifier, and for examples where CaLiGraph and the given classifier disagreed, the human showed slight agreement with the classifier and no agreement with CaLiGraph. 

In addition, the annotator assigned each error to one of five causes: missing data in the entity description that comprised the LLM's ability to classify the entity, a missing class membership relation in the KG between the given entity and class (an example of which is shown in Figure \ref{fig:zero_shot_cot_classifier}), an incorrectly asserted class membership relation in the KG between the given entity and class, and an error on the part of the LLM, through either hallucination or misinterpretation of the class definition or entity description.
\begin{table*}[!ht]
    \centering
    \scriptsize
    \begin{tabular}{p{1.7cm}|p{0.5cm}|p{0.5cm}|p{0.5cm}|p{1cm}|p{1cm}|p{1.4cm}|p{1.3cm}|p{1.4cm}|p{1.4cm}}
        KG & $N$ & $FP$ & $FN$ & human-KG $\kappa$ & human-LLM $\kappa$ & missing data & missing relation & incorrect relation & incorrect reasoning \\ \hline
Wikidata & 136 & 46 & 90 & \textbf{0.243} & -0.241 & 34 (25.0\%) & 15 (11.0\%) & 33 (24.3\%) & 54 (39.7\%) \\
CaLiGraph & 77 & 27 & 50 & -0.295 & \textbf{0.198} & 28 (36.4\%) & 19 (24.7\%) & 20 (26.0\%) & 10 (13.0\%) \\ \hline
& 213 & 73 & 140 & & & 62 (29.1\%) & 34 (16.0\%) & 53 (24.9\%) & 64 (30.0\%)
    \end{tabular}
    \vspace*{2mm}
    \caption{Summary of the analysis of classification errors by gpt-4-0125-preview.}
    \label{tab:summary_error_analysis}
\end{table*}
We assert that these results support another finding: \textbf{classifiers can detect missing or incorrect relations} in KGs ($Q_2$). The error analysis showed that in instances where the classifier using gpt-4-0125-preview was in disagreement with the KG, 40.9\% of errors were due to the knowledge graphs, with 16.0\% due to missing relations and 24.9\% due to incorrect relations. 29.1\% of the errors could be ascribed to missing or insufficient data in the entity description, which may have had a negative impact on classifier performance. This is attributed primarily to one of two reasons: for CaLiGraph, RDF verbalizations missed relevant information about entities due to the omission of relevant triples in the set produced by the SPARQL DESCRIBE queries; and for Wikidata, some entities had descriptions that were simply the label assigned to the entity. We plan to address these shortcomings in future versions of the evaluation datasets. These results suggest that, \textit{pâce} efforts focused on using KGs to mitigate knowledge gaps and hallucinations in LLMs, LLMs may have a corresponding role to play in mitigating knowledge gaps and errors in KGs. 

\section{Discussion}
\begin{description}
\item[Contributions] The principal contributions of this work are 1) a formal approach to the design of a neurosymbolic knowledge engineering workflow integrating KGs and LLMs, and 2) experimental evidence that this method can assist knowledge engineers in addressing the correctness and completeness of KGs, potentially reducing the effort involved in knowledge acquisition and elicitation.
\item[Limitations]
Challenges with the use of LLMs include the cost of API calls to proprietary LLMs and the speed of processing tasks with such resource-intensive systems. Our results show the potential for open source, locally deployed LLMs to address the first problem; we expect that sampling approaches, frequently used in other approaches to KG refinement in large-scale KGs, can help address the second. The human evaluation for error analysis could be improved through the use of crowd-sourcing to expand the number of reviewers (allowing much larger sets of rationales and classification decisions to be evaluated), by evaluating the true positives and true negatives produced by the classifier, and by evaluating the soundness of rationales and faithfulness of classification to the given rationales. The potential impact of one or more of the LLMs having processed the Wikidata and CaLiGraph data during pre-training was not considered in the analysis.  The question of whether the use of gpt-4-1106-preview to generate RDF verbalizations in the CaLiGraph experiments approach to verbalization introduced bias relative to the other LLMs is yet to be addressed. Finally, this work is limited to the evaluation of class membership relations in a KG, and evaluated against KGs that are domain-general and either crowdsourced (Wikidata) or automatically generated from crowdsourced content (CaLiGraph). To support use against KG refinement challenges faced by domain-specific KGs, such as those developed for life sciences applications \cite{chen2023knowledge}, this needs to be generalized to support the definition of classifiers based on intensional definitions of predicates in natural language.
\item[Future work] We have in this work taken a minimalist approach to the prompt engineering of classifiers, restricting ourselves to a zero-shot chain-of-thought approach. Expanding this to include using temperature sampling \cite{ackley1985learning} for self-consistency \cite{wang2022self} and uncertainty estimation \cite{huang2023look}, mitigating hallucination in rationale generation \cite{ji2023survey}, and addressing faithfulness in rationale generation \cite{turpin2023language,agarwal2024faithfulness} are three other areas for future work, in addition to work on addressing the limitations described above by expanding the number and types of relations considered, and evaluating our approach against domain-specific KGs. 
\end{description}

\section*{Acknowledgements}
This work is partially funded by the Dutch Research Council (NWO) through grant MVI.19.032. The authors wish to thank Filip Ilievski, Jan-Christoph Kalo, Xue Li, Fina Polat, Thivyan Thanapalasingam, and Lise Stork  for discussions and suggestions that have been invaluable in refining this work. We would also like to thank the anonymous reviewers for their insightful comments and suggestions, which have been invaluable in refining our work.

\bibliographystyle{splncs04}
\bibliography{bibliography}

\end{document}